\documentclass[runningheads]{llncs}
\usepackage{amsmath}
\usepackage[T1]{fontenc}
\usepackage{graphicx}
\usepackage[linesnumbered,ruled,vlined]{algorithm2e}
\usepackage{hyperref}
\usepackage{xcolor}
\usepackage{booktabs}
\usepackage{enumitem}
\usepackage{array}
\usepackage{amsfonts}
\usepackage{dsfont}
\usepackage{amssymb}
\usepackage{orcidlink}
\usepackage{multirow}
\usepackage{tabularx}
\usepackage[misc]{ifsym}
\usepackage{booktabs}

\usepackage{array}
\SetKwInput{KwInput}{Input}                
\SetKwInput{KwOutput}{Output} 
\newcommand{\etal}[0]{{\textit{et al.}\,}}
\DontPrintSemicolon
\usepackage{algpseudocode} 
\usepackage{academicons}

\begin{document}
\title{Table Detection with Active Learning}
\author{Somraj Gautam \textsuperscript{(\Letter)}  \orcidlink{0009-0004-5648-6772} \and
Nachiketa Purohit  \orcidlink{0009-0007-1620-7793} \and
Gaurav Harit \orcidlink{0000-0001-7943-0123}}
\authorrunning{Gautam et al.}
%
\institute{Indian Institute of Technology Jodhpur\\ Rajasthan, India\\
\email{\{gautam.8, m23csa016, gharit\}@iitj.ac.in}}

\maketitle              
\begin{abstract}
Efficient data annotation remains a critical challenge in machine learning, particularly for object detection tasks requiring extensive labeled data. Active learning (AL) has emerged as a promising solution to minimize annotation costs by selecting the most informative samples. While traditional AL approaches primarily rely on uncertainty-based selection, recent advances suggest that incorporating diversity-based strategies can enhance sampling efficiency in object detection tasks. 
Our approach ensures the selection of representative examples that improve model generalization. We evaluate our method on two benchmark datasets (TableBank-LaTeX, TableBank-Word) using state-of-the-art table detection architectures, CascadeTabNet and YOLOv9. Our results demonstrate that AL-based example selection significantly outperforms random sampling, reducing annotation effort given a limited budget while maintaining comparable performance to fully supervised models.
Our method achieves higher mAP scores within the same annotation budget. 

\end{abstract}
\section{Introduction}
The rapid advancement of machine learning technologies has significantly improved document image analysis. However, the success of these technologies, particularly in object detection tasks, heavily relies on large volumes of annotated data \cite{su2012crowdsourcing}, \cite{russakovsky2015imagenet}, \cite{papadopoulos2017extreme}. This dependency presents a significant challenge, as manual annotation is both time-consuming and expensive. Driven by this fact, researchers have developed methods to choose the most valuable samples in the dataset for labeling, a process called active learning \cite{settles2012active}. Generally, this involves creating a scoring function \cite{lin2024exploring} that assesses the uncertainty in network predictions, prioritizing the labeling of examples where the network's predictions are most uncertain.

Table detection in document images is essential for information extraction, enabling automated systems to identify and interpret structured data within scanned or digital documents. Detecting tables poses unique challenges due to variations in table formats, styles, and layouts in different documents. High-quality bounding box annotations are essential for training robust detection models, but generating such annotations is costly and labor-intensive. The annotation process typically involves marking precise table boundaries, which requires careful manual effort and, in some cases, domain knowledge to distinguish tables from other layout elements. This process is not only labour-intensive but also expensive, as it demands skilled annotators familiar with the intricate layouts and semantics of tables. Consequently, the high cost of labeling impacts the availability of large, well-annotated datasets needed for training table detection models, limiting the performance and generalizability of these models across varied document types.

Active Learning (AL) has emerged as a promising approach to address these challenges. Unlike conventional machine learning techniques that rely on large annotated datasets, AL aims to optimize the learning process by selecting the most informative samples for annotation. Although AL has been successfully applied to various object detection tasks \cite{roy2018deep}, \cite{haussmann2020scalable}, \cite{li2021deep}, \cite{kao2019localization}, \cite{brust2018active}, \cite{yu2022consistency}, \cite{yuan2021multiple}, \cite{choi2021active}, \cite{yang2015multi}, its application to particularly table detection remains unexplored. While active learning has been explored in document layout analysis—for example, \cite{shen2020olala} applies active learning to multi-class layout detection, including tables- our work specifically focuses on single-class table detection. To the best of our knowledge, no prior work has addressed the unique challenges associated with applying active learning exclusively to table detection. We hypothesize that this focused approach not only reduces labeling effort but also enhances model performance by targeting the most uncertain or ambiguous table instances. Two primary AL strategies have been commonly explored in the literature: (1) Uncertainty-based methods \cite{joshi2009multi}, \cite{lewis1994heterogeneous}, \cite{lewis1995sequential}, \cite{roth2006margin}, \cite{learningsynthesis}, focus on selecting samples where the model predictions exhibit the highest uncertainty. This includes cases where the model has the least confidence in its predictions, shows a high entropy in class probabilities, or has a minimal margin between the top predicted classes. (2) Diversity-based methods \cite{agarwal2020contextual}, \cite{bilgic2009link}, \cite{elhamifar2013convex}, \cite{guo2010active}, \cite{luo2013latent}, \cite{mac2014hierarchical}, \cite{sener2017active}, \cite{wang2016cost}, \cite{yang2015multi} aim to choose samples that best represent the overall data distribution by minimizing similarities between selected samples, using feature representations and identifying a subset using a distance metric \cite{sener2017active}.

Applying Active Learning (AL) to table detection introduces unique challenges distinct from general object detection tasks. Unlike multiclass object detection, where uncertainty can be measured across different categories, table detection involves a single class, making uncertainty estimation less informative. This lack of class diversity poses difficulties for traditional uncertainty-based strategies, which typically rely on class probability distributions. Diversity-based methods pose a difficulty in defining optimal selection criteria because intraclass variations are limited.  This gap in research requires novel strategies that balance informativeness and diversity to optimize example selection for annotation and training purposes. 

Our work identifies the challenges of uncertainty-based sampling when applied to table detection and introduces a new diversity-based sampling strategy tailored to this task. Furthermore, we propose a new active learning strategy that leverages model predictions for spatial and structural information to identify hard examples for which the model struggles in its predictions. These include discrepancies between segmentation and detection masks, overlapping table regions, and multi-table configurations. By incorporating model confidence scores, our approach ensures that selected samples are both informative and representative.

To validate our method, we conduct extensive experiments on two widely recognized benchmark datasets: TableBank-LaTeX and TableBank-Word. We employ state-of-the-art detection architectures, namely YOLOv9 and CascadeTabNet, to evaluate the effectiveness of our diversity-based and hybrid-based active learning strategy. Our results demonstrate significant improvements over random sampling and traditional uncertainty-based approaches, achieving higher mean Average Precision (mAP) scores within the same annotation budget.

Our major 2-fold contributions are as follows:
\begin{itemize}
    \item \textbf{Novel active learning strategies for table detection:} We develop new selection strategies designed specifically for single-class table detection, addressing the limitations of existing AL methods. We use model prediction uncertainty and prediction ambiguities to formulate the sampling strategies. 
    \item \textbf{Empirical Validation with Benchmark Datasets:} Extensive experiments on TableBank-LaTeX and TableBank-Word using YOLOv9 and CascadeTabNet demonstrate that our active learning strategies significantly reduce annotation costs while maintaining comparable or superior performance to fully supervised models.
\end{itemize}

We begin by reviewing the state-of-the-art table detection methodologies, identifying their limitations, and exploring the potential benefits of integrating Active Learning (AL) to optimize annotation efforts. We then introduce our proposed AL framework for table detection, detailing its sample selection criteria that combine uncertainty-based and diversity-based strategies to ensure a more informative and representative selection process.

\section{Related Work}
Table classification and detection are widely explored topics in document image analysis \cite{prasad2020cascadetabnet}, \cite{liu2024improving}, \cite{gao2019icdar}, \cite{li2020TableBank}. In this work, we explore existing state-of-the-art architectures $-$ single-stage detector (YOLO) and multistage detector (Cascade-mask RCNN) for their superior accuracy, ease of use, and speed. However, these methods use full supervision in terms of ground-truth bounding boxes for object detection. The cost of fully annotating images is high, and so weak supervision has received attention over the last few years as it needs only image labels for training \cite{bilen2016weakly}. In this paper, we focus on a related paradigm called active learning \cite{settles2009active}, which provides state-of-the-art performance with minimum supervision. Active learning has been extensively studied over the past decade \cite{garcia2023ten}, particularly in image classification tasks. 

Roy \etal \cite{roy2018deep} classify Active Learning methods into two categories: (1) black-box methods and (2) white-box methods. The key difference lies in their level of access to the model architecture. Black-box approaches operate without internal model insights, relying solely on softmax-derived confidence scores for sample selection. In contrast, white-box methods utilize internal model parameters, making them inherently architecture dependent. 

There have been limited applications of active learning for table detection. 
Dong et al \cite{dong2019tablesense} have used active learning to label data in iterations, where they define an effective uncertainty metric as the key to selecting the least confident sheets in spreadsheets to label in the next iteration. 

\subsection{Active Learning Sampling Strategies}
Settles \cite{settles2009active} provided a seminal review of active learning techniques, categorizing sampling approaches into uncertainty sampling, query-by-committee, and density-based methods. Researchers have adapted these strategies to address unique challenges in the context of document image analysis.
Saifullah \etal \cite{saifullah2023analyzing} introduced an active learning framework for document image classification, demonstrating that selective sampling can reduce annotation efforts by up to 60\% while maintaining high accuracies.

\subsubsection{Uncertainty-based}~\\
In uncertainty sampling, which is among the most popular approaches \cite{settles2012active}, the active learner sequentially queries the label of instances for which its current prediction is maximally uncertain. Naguyen \etal \cite{ALentropy} have proposed using entropy as a measure of uncertainty in active learning. 
For an image $i$, consider that we have $b_{it}$ set bounding boxes corresponding to each table $t$. We define the entropy of each box using the binary entropy formula. The query function selects the image with the most uncertain table:
\[
\operatorname*{argmax}_i \max_t \sum_{y \in b_{it}} \left[ 
  -p(y|x)\log p(y|x) - (1 - p(y|x))\log (1 - p(y|x)) 
\right]
\]

Here, the term in brackets denotes the binary entropy, which captures the model’s uncertainty about each bounding box. 

\subsubsection{Diversity-based}~\\
Diversity-based sampling has emerged as a complementary approach to uncertainty sampling, particularly in computer vision tasks. \cite{nguyen2004active} introduced clustering-based methods to facilitate diverse representation in the selected samples. Similarly, \cite{sener2017active} proposed selection of a core set, which is a carefully chosen set of unlabeled examples that are the most diverse and representative to maximize the coverage of the feature space while minimizing redundancy. These approaches have shown promising results in the image classification task but have seen limited application for detection problems.

\subsubsection{Combination of diversity-based and uncertainty-based approaches}~\\
Recent research has demonstrated the benefits of combining uncertainty and diversity criteria. Wu \etal \cite{wu2022entropy} proposed active learning for object detection with a two-stage selection process that first identifies uncertain samples and then applies diversity filtering.

\subsection{Confidence score calculation}
The confidence score in object detection models typically involves softmax and sigmoid activation functions. We compute:
\begin{itemize}
    \item Predicted class probability.
    \item Intersection over Union (IoU) for bounding box accuracy.
    \[Confidence = P(class) * IoU\]
\end{itemize}
Here $P(class)$ is the predicted class probability using softmax and sigmoid activation functions.

\begin{table}
\centering
\caption{Comparative Analysis of Confidence Score Approaches}
\label{table1}
\begin{tabular}{|m{2.4cm}|m{4.7cm}|m{4.7cm}|}
\hline
\multicolumn{1}{|c|}{\textbf{Architecture}} & 
\multicolumn{1}{c|}{\textbf{YOLOv9 \cite{wang2024yolov9learningwantlearn}}} & 
\multicolumn{1}{c|}{\textbf{CascadeTabNet \cite{prasad2020cascadetabnet}}} \\ \hline

Confidence Score Computation & 
Sigmoid-based activation with objectness-class probability multiplication. & 
Softmax-based class probability estimation with multi-stage refinement. \\ \hline

Characteristics &  
\begin{minipage}[t]{\linewidth}
    \begin{itemize}[nosep,leftmargin=*,label=\textbullet]
    \item Single-stage detection.
    \item Direct confidence calculation.
    \vspace{2pt}
\end{itemize}
\end{minipage} & 
\begin{minipage}[t]{\linewidth}
    \begin{itemize}[nosep,leftmargin=*,label=\textbullet]
        \item Multi-stage refinement.
        \item IoU-based confidence scoring.
        \vspace{2pt}
    \end{itemize}
\end{minipage} \\ \hline

Advantages & 
\begin{minipage}[t]{\linewidth}
    \begin{itemize}[nosep,leftmargin=*,label=\textbullet]
        \item High-speed inference.
        \item End-to-end optimization.
        \vspace{2pt}
    \end{itemize}
\end{minipage} & 
\begin{minipage}[t]{\linewidth}
    \begin{itemize}[nosep,leftmargin=*,label=\textbullet]
        \item Higher detection accuracy.
        \item Improved classification precision.
        \vspace{2pt}
    \end{itemize}
\end{minipage} \\ \hline

Limitations & 
\begin{minipage}[t]{\linewidth}
    \begin{itemize}[nosep,leftmargin=*,label=\textbullet]
        \item Struggles with overlapping objects.
        \item Grid constraints in small object detection.
        \item Lower precision in complex scenes.
        \vspace{2pt}
    \end{itemize}
\end{minipage} & 
\begin{minipage}[t]{\linewidth}
    \begin{itemize}[nosep,leftmargin=*,label=\textbullet]
        \item Increased computational overhead.
        \item Higher training complexity.
        \item Slower inference speed.
        \vspace{2pt}
    \end{itemize}
\end{minipage} \\ \hline

\end{tabular}
\end{table}

The comparative analysis of object detection architectures reveals distinct approaches to confidence score calculation. As shown in Table \ref{table1}, YOLOv9 \cite{wang2024yolov9learningwantlearn} utilizes sigmoid activation with direct objectness and class probability multiplication, while CascadeTabNet \cite{prasad2020cascadetabnet} employs softmax for class probability estimation with multi-stage refinement. These architectural variations significantly impact the models' ability to predict object presence and classification with varying degrees of precision and computational efficiency.

\section{Proposed Active Learning Methodology}
\begin{figure}[t]
\centering
\includegraphics[width=\textwidth]{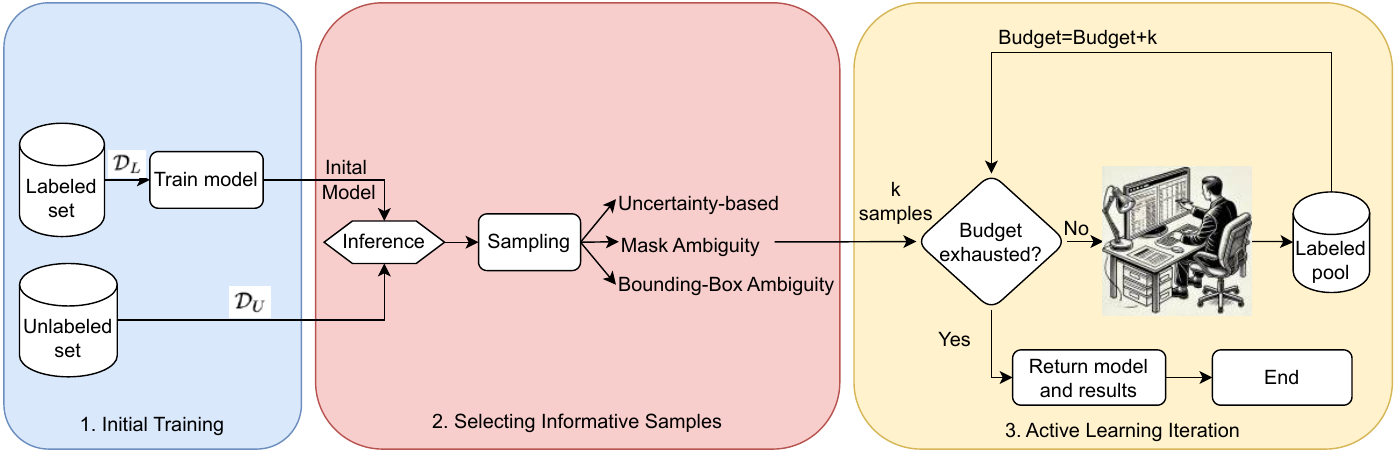}
\caption{Our proposed Active learning Framework, k, is the number of samples we want to increase in every iteration in our budget.} \label{Fig1}
\end{figure}

In this section, we first define our proposed Active Learning framework, followed by an uncertainty-based approach for table detection in document images.  
Figure \ref{Fig1} and Algorithm~\ref{alg:table_detection_active_learning} provide a high-level overview of our Active Learning framework. We make use of 4 sampling strategies based on uncertainties in model prediction (Section~\ref{sec:sampling_prediction_uncertainty}) and ambiguities in model predictions (Section~\ref{sec:sampling_model_ambiguities}). Our framework utilizes a white-box approach where we trained and evaluated two state-of-the-art models—YOLOv9 and CascadeTabNet on the TableBank
dataset. 

\begin{algorithm}[ht]
\caption{Active Learning for Table Detection}
\label{alg:table_detection_active_learning}
\SetKwInOut{Input}{Input}
\SetKwInOut{Output}{Output}
\SetKwFunction{TrainModel}{TrainModel}
\SetKwFunction{RunInference}{RunInference}
\SetKwFunction{ApplySampling}{ApplySampling}
\SetKwFunction{SelectSubset}{SelectSubset}
\SetKwFunction{AnnotateData}{AnnotateData}
\SetKwFunction{EvaluateModel}{EvaluateModel}

\Input{
    Training dataset $\mathcal{D} = \{x_1, x_2, \dots, x_n\}$,
    Total annotation budget $B$, Initial labeled dataset size $K$,
    Step size for budget increment $k$, Starting budget $\epsilon$.
}
\Output{
    Optimized model $\mathcal{M}^{*}$, Performance metrics across iterations
}
$\mathcal{D}_L \gets \{x_1, x_2, \dots, x_K\}$ \Comment {The set of labelled examples available initially}\\
$\mathcal{D}_U \gets \mathcal{D} \setminus \mathcal{D}_L$ \Comment{The set of remaining unlabelled examples}\\
$B_{\text{remaining}} \gets B - K$ \Comment{Remaining annotation budget available for utilization}\\
$b \gets \epsilon$ \Comment{Starting budget}\\
$\mathcal{M}_0 \gets \TrainModel(\mathcal{D}_L)$\\
$P \gets \RunInference(\mathcal{M}_0, \mathcal{D}_U)$ \\
$L \gets \ApplySampling(P, \mathcal{D}_U)$ \Comment{Active learning methods used here} \\
$\mathcal{D}_{L}^{\text{new}} \gets \phi$ \\

\While{$b \leq B_{\text{remaining}}$}{
    $S_L \gets \SelectSubset(L, \min(k, B_{\text{remaining}}))$\\ 
    $S_L \gets \AnnotateData(S_L)$ \\
    $\mathcal{D}_{L}^{\text{new}} \gets \mathcal{D}_{L}^{\text{new}} \cup S_L$ \\
    $\mathcal{M} \gets \TrainModel(\mathcal{D}_{L}^{\text{new}})$ \Comment{Continue training from the
previous state.}\\
    $\text{metrics} \gets \EvaluateModel(\mathcal{M})$\\
    $b \gets b + |S_L|$\\
    $L \gets L \setminus S_L$
}
\Return{$\mathcal{M}^{*} = \mathcal{M}, \text{metrics}$}
\end{algorithm}

\subsection{Sampling hard examples using prediction uncertainty}
\label{sec:sampling_prediction_uncertainty}
Following an active learning paradigm, we first train an initial model \( f_0^\theta \) on a randomly selected subset \( D_0^L \subset D \) of the training data $D$. This model is then used to assess the difficulty level of examples in the training set by computing confidence scores \( c_i = f_0^\theta(d_i) \) of the predicted bounding box for tables present in each image \( d_i \in D \). To identify informative samples that could improve model performance, we employ our proposed confidence binning method, which can reduce the redundancy of similar confidence scores and encourage the selection of examples with low confidence scores for the detected bounding boxes. 
Our approach creates confidence bins and segregates the document page into the bins using their confidence values. We denote the ranges (bins) for the confidence values as $R_j$. 
\[ H = \bigcup_{j=1}^k { \textsc{Sample}(R_j, r_j) } \]
Examples in a given confidence bin are subjected to sampling with a sampling rate specific to that confidence bin. The sampling process creates an informative sample set ($H$) defined as a union of the samples obtained by applying the sampling function \textsc{Sample} to each confidence range $R_j$ with the corresponding sampling rate $r_j$.

As an illustration, let's say we create confidence bins, such as 40-50\%, 50-60\%, . . 60-95\%. The lower the confidence, the more likely it is an uncertain sample, and we prioritize selecting from those bins. This ensures a diverse and balanced selection of informative samples $H$.
The final model \( f_{H}^\theta \) is trained on the selected informative samples, and then these selected informative samples are given to humans for annotation. This selective approach and human-in-the-loop annotations significantly reduce the annotation workload.

We compute the sampling rate $r_j$ (proportion of examples selected from the bin $R_j$) for the confidence bin $R_j$ as follows: 
\[r_j = \max\left(0, 100 - (R_{j_{\text{low}}} - R_{\text{min}})\right)\]
where $R_{j_{\text{low}}}$ is the lower bound of the confidence range $R_j$ and $R_{\text{min}}$ is the minimum confidence score (40 for the example bins given above). 

The effect of the sampling strategy is that the proportion of selected examples decreases linearly as the confidence score range increases, ensuring higher selection rates for low-confidence samples and progressively fewer selections for higher-confidence samples. The approach balances exploration by including more of lower confidence samples and also some high confidence samples.

\begin{figure}
\includegraphics[width=0.45\textwidth]{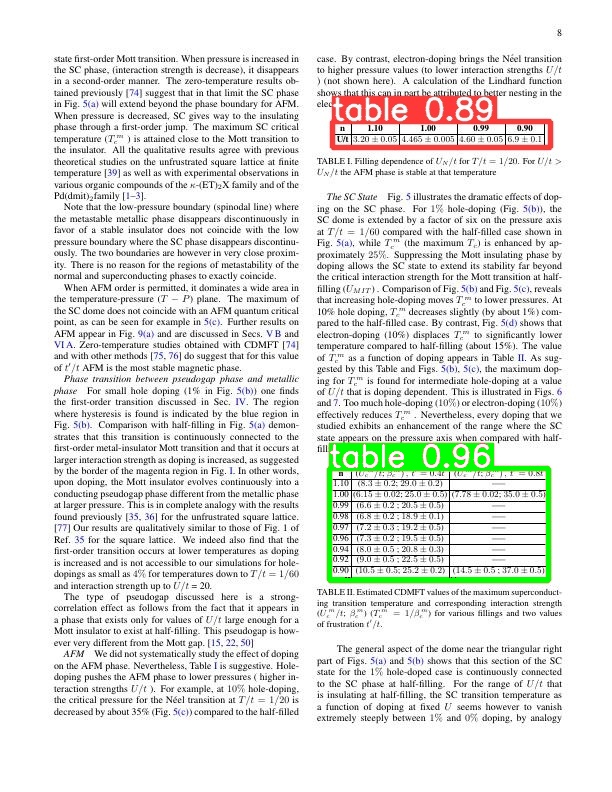}
\vline
\includegraphics[width=0.45\textwidth]{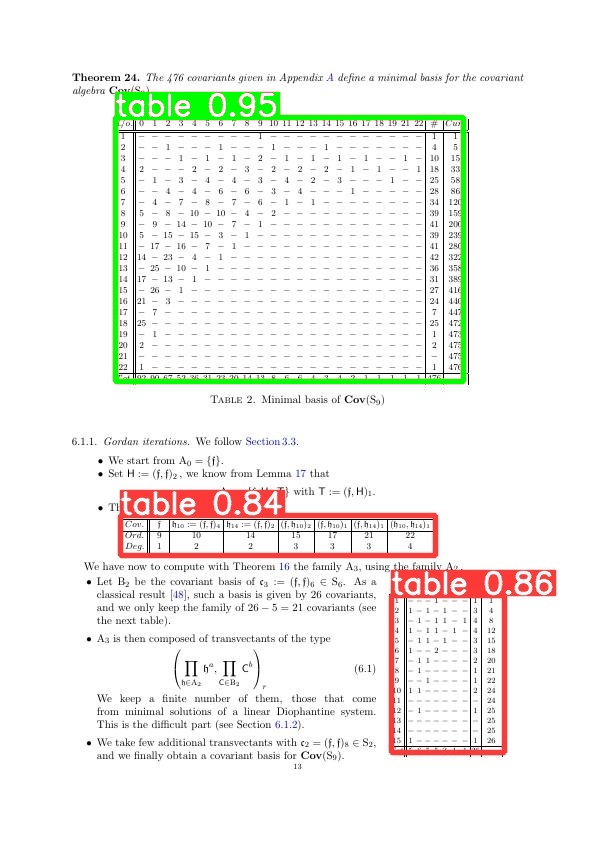}
\caption{Figure shows examples of documents with multiple images. The confidence scores are predicted for each table distinctly. We take the average of all the confidence scores for predictions in an image.} \label{Fig2}
\end{figure}

Many images in our dataset contain more than one table. As shown in Figure~\ref{Fig2}, when multiple tables are present in a single document, we compute a confidence score for each detected table and then average these scores to obtain the overall detection confidence for the image.   

\subsection{Sampling hard examples using ambiguities in model predictions}
\label{sec:sampling_model_ambiguities}
In addition to the prediction confidence, we also identify other cues that indicate that the model has struggled while making its predictions. We formulate 3 such measures to identify and select the hard examples for training. 

\begin{figure}
\centering
\includegraphics[width=0.9\textwidth]{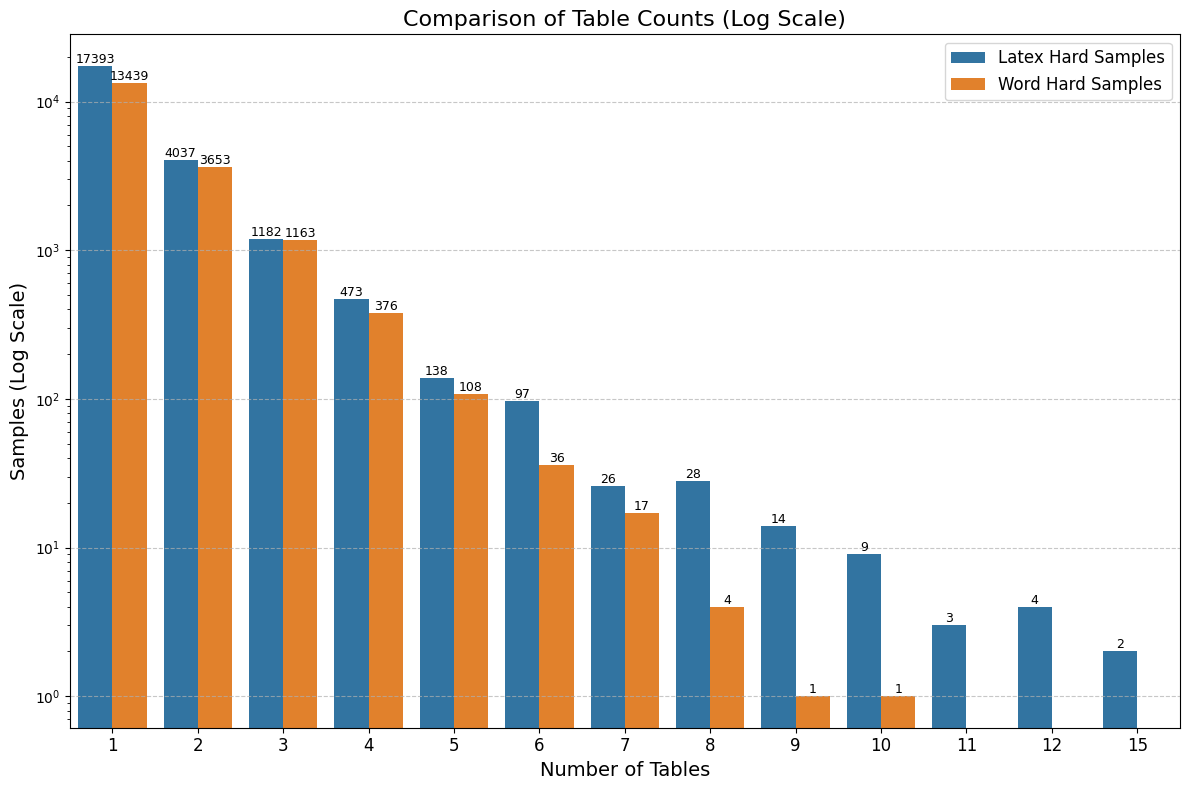}
\caption{Distribution of hard samples across multi-table images which we used in Table Count strategy for TableBank-LaTeX and TableBank-Word datasets.} \label{Fig3}
\end{figure}

\begin{enumerate}
\item We consider that when a model produces multiple detections that overlap with one another, it reflects the model's confusion in delineating closely located tables. We capture the performance gap by formulating a measure that takes a higher value if the predicted bounding boxes are overlapping. We formulate this \textbf{bounding-box ambiguity (BBA) score} as the number of detections that have significant (greater than a threshold) overlap with another one. This is then normalized by the number of detections in the image to give a value in the range 0 to 1, as follows: 
\[\text{BBA score} = \frac{\sum\limits_{i=1}^{N_d} \mathds{1} \left( \max\limits_{j \neq i} \text{IoU}(d_i, d_j) > T_{\text{IoU}} \right)}{N_d}\]
where:
\begin{itemize}
    \item \( N_d \) is the total number of detected bounding boxes.
    \item \( d_i \) represents the \( i \)-th detected bounding box.
    \item \( T_{\text{IoU}} \) is a predefined IoU threshold.
    \item \( \mathds{1}(\cdot) \) is an indicator function that returns 1 if the condition inside is true; otherwise, 0.
    \item \( \max\limits_{j \neq i} \text{IoU}(d_i, d_j) \) finds the highest IoU value between detection \( d_i \) and any other detection \( d_j \).
\end{itemize}

\item We measure the deviation between the segmentation mask and the detection mask. We formulate the \textbf{mask ambiguity (MA) score } as $(1-\text{IoU})$ where IoU is computed between the detection mask and the segmentation mask. We do not need to compute this score separately for each table in the page, since we are computing IoU between all the bounding box occupied regions and the segmentation mask occupied regions. 

\item  The third sampling strategy considers images with more than 1 table as hard examples that can be included for sampling. We assign images with more tables a higher sampling probability proportionate to the number of tables. 
Images that have low prediction confidence are given to human annotators. Amongst the annotated images, we identify images with more than one table and assign a higher sampling probability to them. We call this strategy as \textbf{table count (TC)} based strategy. Figure \ref{Fig3} shows the distribution of multi-table images across both datasets, demonstrating that TableBank-LaTeX contains a higher proportion of documents with multiple tables, which validates our table count-based sampling strategy.

\end{enumerate}

\section{Results and Analysis}
This section presents the experimental findings of our proposed Active Learning framework for table detection. We evaluate and compare various active learning strategies across two benchmark datasets from the TableBank corpus.

\subsection{Dataset Description}
\textbf{TableBank-LaTeX}: This dataset consists of document images extracted from LaTeX sources, primarily scientific papers and academic publications. The tables in this dataset are characterized by:
\begin{itemize}
    \item Structured, grid-like layouts with clear boundaries.
    \item Consistent formatting typical of academic publications.
    \item Predominantly black-and-white text with minimal graphics.
    \item Higher uniformity in tables due to LaTeX's standardized formatting.
\end{itemize}
\textbf{TableBank-Word}: This dataset comprises images from Microsoft Word documents, including business reports, forms, and various document types. The tables exhibit:
\begin{itemize}
    \item More diverse and irregular table layouts.
    \item Varied formatting styles, fonts, and colors.
    \item Complex table structures with merged cells and nested elements.
    \item Greater variability in table appearance due to user-defined formatting.
\end{itemize} 
The statistics of the train, validation, and test splits used in our experiments are summarized in Table~\ref{tab:dataset-stats}.
The significant difference in dataset sizes (TableBank-LaTeX being larger) also allows us to assess how our active learning strategies perform across different data availability scenarios.

\begin{table}[h]
\centering
\caption{Statistics of TableBank-LaTeX and TableBank-Word splits based on number of images.}
\begin{tabular}{|l|c|c|c|}
\hline
\textbf{Source} & \textbf{Train} & \textbf{Validation} & \textbf{Test}\\
\hline
LaTeX & 187,199 & 7,265 & 5,719 \\ \hline
Word & 73,383 & 2,735 & 2,281 \\
\hline
\end{tabular}
\label{tab:dataset-stats}
\end{table}
\noindent \textbf{Experimental Setup}: The assessment is conducted using two state-of-the-art table detection architectures: YOLOv9 and CascadeTabNet. We set the uncertainty threshold to 0.95. To ensure a sufficiently large sample set under budget constraints, we set the IoU threshold $T_{IoU}$ to 0.006 for TableBank-Word and 0.004 for TableBank-LaTeX when computing the BBA score.

\begin{table}[t]
\centering
\caption{Result (mAP scores) of CascadeTabNet using our proposed approach.}
\begin{tabular}{|l|lllll!{\vrule width 1.5pt}lllll|}
\hline
\multirow{3}{*}{Type} & \multicolumn{5}{c!{\vrule width 1.5pt}}{TableBank-Word}& \multicolumn{5}{c|}{TableBank-LaTeX}\\ \cline{2-11} 
& \multicolumn{10}{c|}{Budget}        \\ \cline{2-11} 
& \multicolumn{1}{l|}{2k}  & \multicolumn{1}{l|}{4k}  & \multicolumn{1}{l|}{6k}  & \multicolumn{1}{l|}{8k}  & \multicolumn{1}{l!{\vrule width 1.5pt}}{10k} & \multicolumn{1}{l|}{2k}  & \multicolumn{1}{l|}{4k}  & \multicolumn{1}{l|}{6k}  & \multicolumn{1}{l|}{8k}  & \multicolumn{1}{l|}{10k} \\ \hline
Random      & \multicolumn{1}{l|}{82.2}& \multicolumn{1}{l|}{85.1}& \multicolumn{1}{l|}{86.8}& \multicolumn{1}{l|}{87.8}& \multicolumn{1}{l!{\vrule width 1.5pt}}{88.8}& \multicolumn{1}{c|}{\textbf{86.7}} & \multicolumn{1}{c|}{87.2}& \multicolumn{1}{c|}{90.6}& \multicolumn{1}{c|}{91.8}& 91.2 \\ \hline
Prediction Uncertainty  & \multicolumn{1}{l|}{\textbf{83.1}} & \multicolumn{1}{l|}{\textbf{87.3}} & \multicolumn{1}{l|}{88.1}& \multicolumn{1}{l|}{\textbf{89.4}} & \multicolumn{1}{l!{\vrule width 1.5pt}}{89.8}& \multicolumn{1}{c|}{80.6}& \multicolumn{1}{c|}{\textbf{90.7}} & \multicolumn{1}{c|}{\textbf{91.6}} & \multicolumn{1}{c|}{92.4}& 91.2 \\ \hline
Mask Ambiguity    & \multicolumn{1}{l|}{52.1}& \multicolumn{1}{l|}{86.6}& \multicolumn{1}{l|}{\textbf{88.4}} & \multicolumn{1}{l|}{89.3}& \multicolumn{1}{l!{\vrule width 1.5pt}}{89.5}& \multicolumn{1}{c|}{83.9}& \multicolumn{1}{c|}{88.9}& \multicolumn{1}{c|}{\textbf{91.6}} & \multicolumn{1}{c|}{92.4}& 92.9  \\ \hline
Bounding-Box Ambiguity     & \multicolumn{1}{l|}{69.7}& \multicolumn{1}{l|}{84.2}& \multicolumn{1}{l|}{87.4}& \multicolumn{1}{l|}{88.9}& \multicolumn{1}{l!{\vrule width 1.5pt}}{\textbf{90.3}} & \multicolumn{1}{c|}{80.3}& \multicolumn{1}{c|}{88.0}& \multicolumn{1}{c|}{90.4}& \multicolumn{1}{c|}{91.3}& 92.2 \\ \hline
Table Count  & \multicolumn{1}{c|}{80.3}& \multicolumn{1}{c|}{85.5}& \multicolumn{1}{c|}{86.9}& \multicolumn{1}{c|}{88.6}& \multicolumn{1}{c!{\vrule width 1.5pt}}{89.3}& \multicolumn{1}{c|}{77.9}& \multicolumn{1}{c|}{89.0}& \multicolumn{1}{c|}{91.3}& \multicolumn{1}{c|}{\textbf{92.5}} & \textbf{92.7} \\ \hline
\end{tabular}
\label{Table2}
\end{table}

\subsection{Performance Comparison}
Tables \ref{Table2} and \ref{Table3} summarize the mAP scores achieved by each sampling strategy across different annotation budgets (2k to 10k samples). Several key observations emerge:

\textbf{CascadeTabNet on TableBank-Word}: The bounding box ambiguity score-based active learning delivers the best performance on TableBank-Word, reaching an mAP of 90.3\% at 10k samples. 

\textbf{CascadeTabNet on TableBank-LaTeX}: The mask ambiguity score-based active learning excels at higher budget levels on TableBank-LaTeX, achieving an mAP of 92.9\% at 10k samples. 

\textbf{YOLOv9 on TableBank-Word}: Bounding-box ambiguity approach consistently performs well in mid-range budgets (6k–8k), reaching an mAP of 91.7\% with 8k samples. At 10k samples, the table count approach achieves the highest performance (91.0\%).

\textbf{YOLOv9 on TableBank-LaTeX}: The table count approach demonstrates the best performance, achieving an mAP of 89.2\% with just 2k samples, significantly outperforming both random sampling (78.5\%) and uncertainty-based sampling (74.1\%). This advantage persists at higher budgets, reaching its peak performance of 95.9\% at 10k samples, the highest performance observed across all configurations.

We also observe that for higher annotation budgets, the performance gaps between random sampling and active learning are closed, mainly because even with random sampling, an adequate number of hard examples are included for labelling. 

\begin{table}[t]
\centering
\caption{Result (mAP scores) of YOLOv9 using our proposed approach}
\begin{tabular}{|l|llllllllll|}
\hline
\multirow{3}{*}{Type} & \multicolumn{5}{c!{\vrule width 1.5pt}}{TableBank-Word} & \multicolumn{5}{c|}{TableBank-LaTeX} \\ \cline{2-11} 
 & \multicolumn{10}{c|}{Budget} \\ \cline{2-11} 
 & \multicolumn{1}{l|}{2k} & \multicolumn{1}{l|}{4k} & \multicolumn{1}{l|}{6k} & \multicolumn{1}{l|}{8k} & \multicolumn{1}{l!{\vrule width 1.5pt}}{10k} & \multicolumn{1}{l|}{2k} & \multicolumn{1}{l|}{4k} & \multicolumn{1}{l|}{6k} & \multicolumn{1}{l|}{8k} & 10k \\ \hline
Random & \multicolumn{1}{l|}{64.6} & \multicolumn{1}{l|}{84.2} & \multicolumn{1}{l|}{88.0} & \multicolumn{1}{l|}{89.6} & \multicolumn{1}{l!{\vrule width 1.5pt}}{88.6} & \multicolumn{1}{l|}{78.5} & \multicolumn{1}{l|}{87.3} & \multicolumn{1}{l|}{88.7} & \multicolumn{1}{l|}{90.1} & 92.1 \\ \hline
Prediction Uncertainty & \multicolumn{1}{l|}{67.3} & \multicolumn{1}{l|}{86.3} & \multicolumn{1}{l|}{89.2} & \multicolumn{1}{l|}{89.2} & \multicolumn{1}{l!{\vrule width 1.5pt}}{90.1} & \multicolumn{1}{l|}{74.1} & \multicolumn{1}{l|}{88.1} & \multicolumn{1}{l|}{89.6} & \multicolumn{1}{l|}{90.4} & 92.8 \\ \hline
Mask Ambiguity & \multicolumn{1}{l|}{66.4} & \multicolumn{1}{l|}{84.6} & \multicolumn{1}{l|}{89.4} & \multicolumn{1}{l|}{89.6} & \multicolumn{1}{l!{\vrule width 1.5pt}}{90.5} & \multicolumn{1}{l|}{74.1} & \multicolumn{1}{l|}{88.4} & \multicolumn{1}{l|}{88.4} & \multicolumn{1}{l|}{89.1} & 91.3 \\ \hline
Bounding-Box Ambiguity & \multicolumn{1}{l|}{\textbf{68.2}} & \multicolumn{1}{l|}{87.9} & \multicolumn{1}{l|}{90.2} & \multicolumn{1}{l|}{\textbf{91.7}} & \multicolumn{1}{l!{\vrule width 1.5pt}}{90.2} & \multicolumn{1}{l|}{76.3} & \multicolumn{1}{l|}{89.1} & \multicolumn{1}{l|}{89.5} & \multicolumn{1}{l|}{90.4} & 91.3 \\ \hline
Table Count & \multicolumn{1}{l|}{67.0} & \multicolumn{1}{l|}{\textbf{89.3}} & \multicolumn{1}{l|}{\textbf{90.7}} & \multicolumn{1}{l|}{90.3} & \multicolumn{1}{l!{\vrule width 1.5pt}}{\textbf{91.0}} & \multicolumn{1}{l|}{\textbf{89.2}} & \multicolumn{1}{l|}{\textbf{90.9}} & \multicolumn{1}{l|}{\textbf{91.9}} & \multicolumn{1}{l|}{\textbf{92.3}} & \textbf{95.9} \\ \hline
\end{tabular}
\label{Table3}
\end{table}

\subsection{Annotation Efficiency}
\begin{figure}
\centering
\includegraphics[width=1\textwidth]{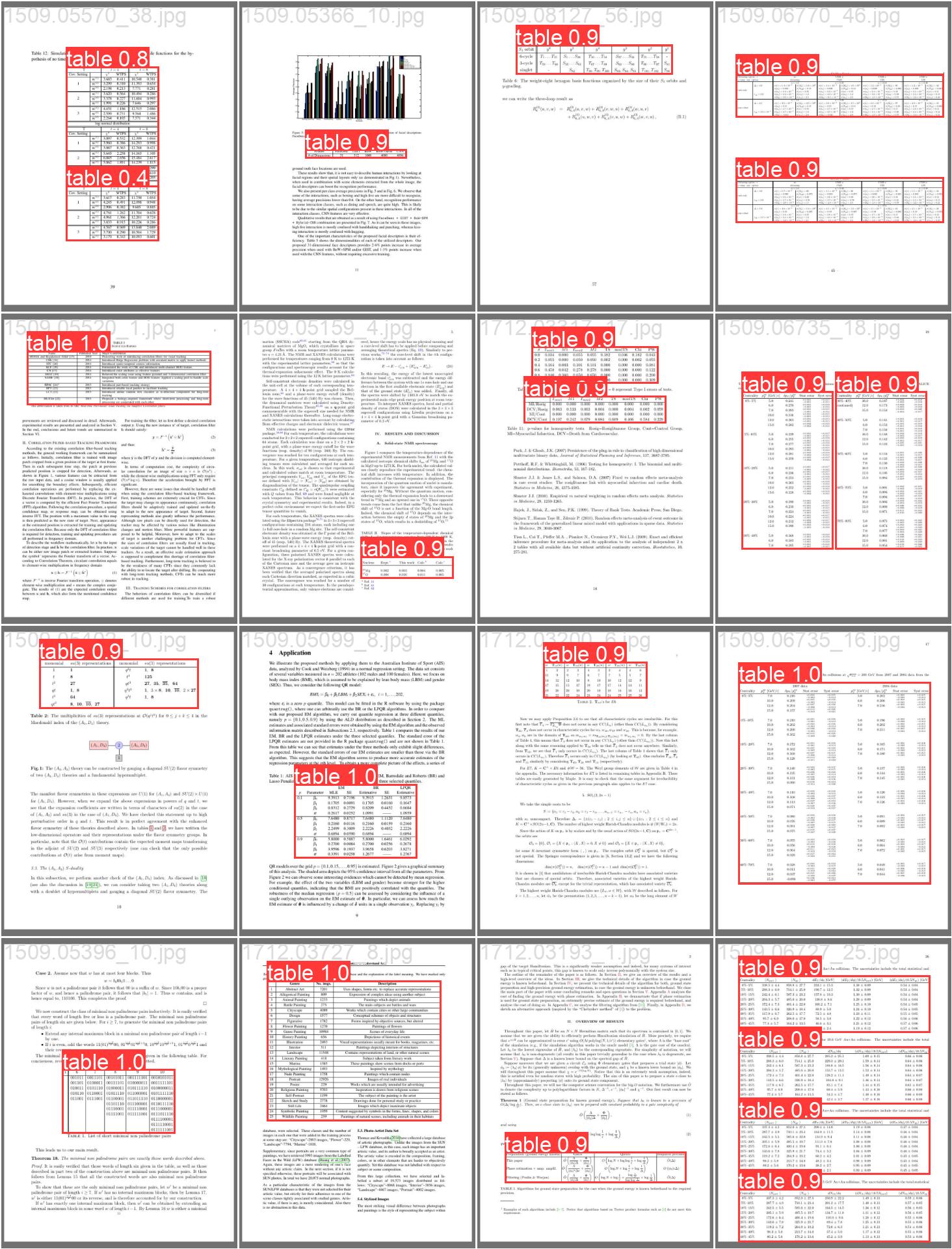}
\caption{Hard samples identified by using active learning on TableBank-Latex dataset.}
\label{Fig6}
\end{figure}

\begin{figure}
\centering
\includegraphics[width=1\textwidth]{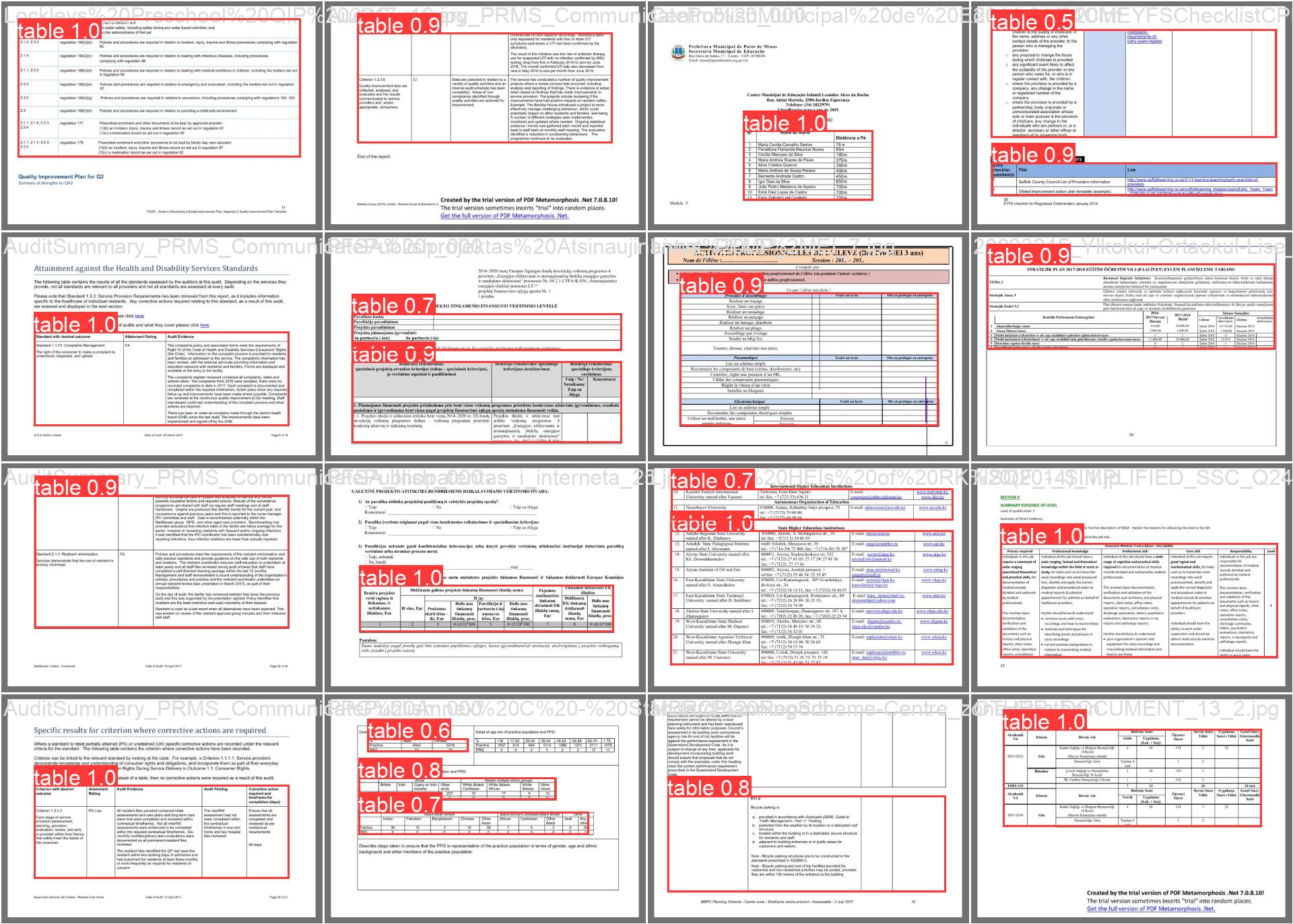}
\caption{Hard samples identified by using active learning on TableBank-Word dataset.}
\label{Fig5}
\end{figure}

\begin{figure}
\centering
\includegraphics[width=1\textwidth]{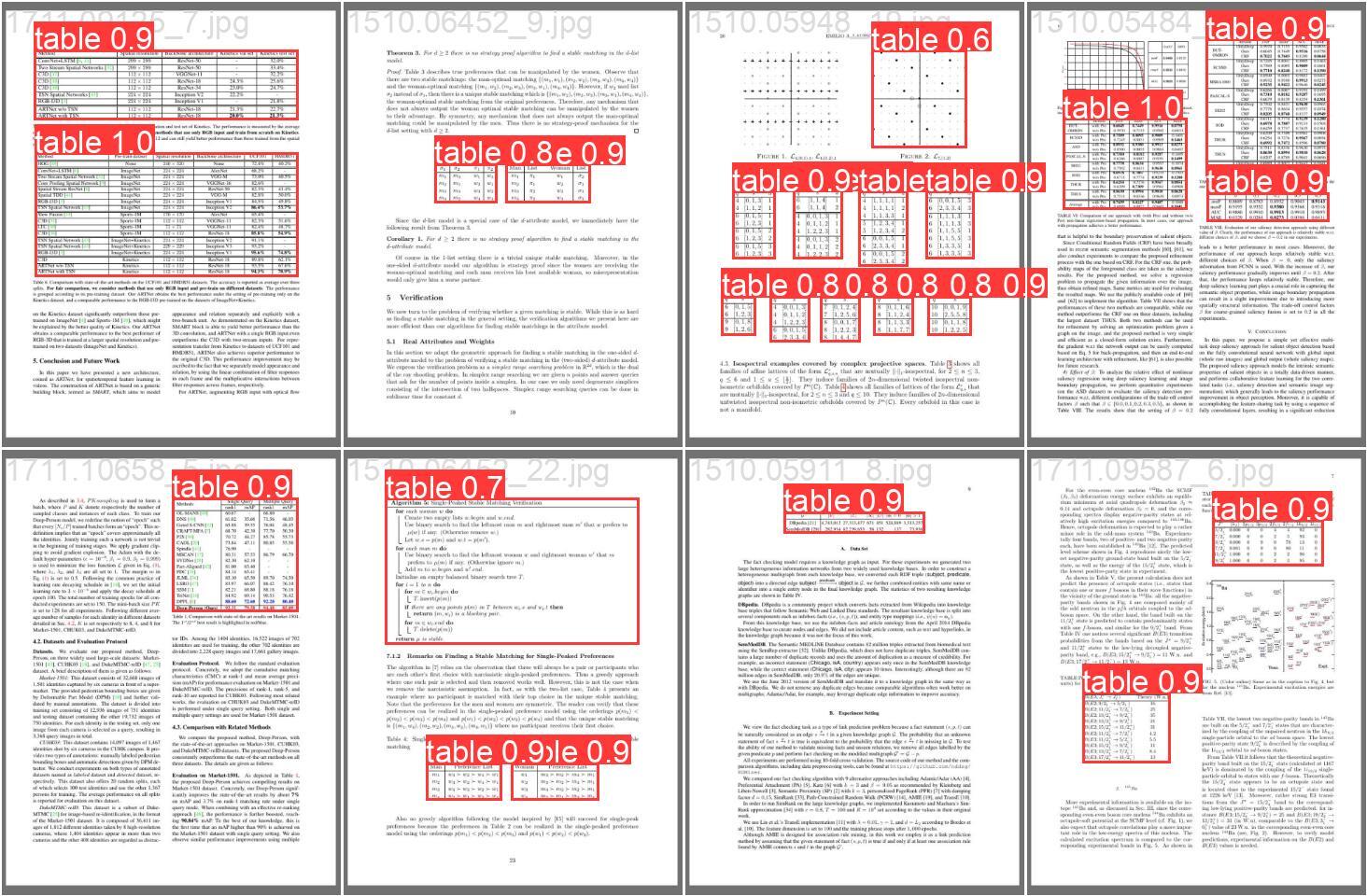}
\caption{More Hard samples of TableBank-LaTeX dataset}
\label{Fig4}
\end{figure}

The advantages of sampling approaches based on ambiguities in model prediction become particularly evident when analyzing performance relative to annotation budgets. For instance, the prediction uncertainty-based approach with CascadeTabNet on TableBank-Word achieves an mAP of 89.4\% in just 8k samples, achieving a 20\% reduction in the annotation with respect to random samples, whereas on TableBank-LaTeX, there is a 40\% reduction in the annotation cost. Referring to Table \ref{Table3}, we observe that the table count approach with YOLOv9 on TableBank-LaTeX achieves an mAP of 89.2\% with just 2k samples—a performance level that random sampling requires 8k samples to match, representing a 75\% reduction in annotation effort with 6k fewer samples annotated. The effectiveness of our approach is demonstrated through qualitative examples shown in Figures \ref{Fig6}, \ref{Fig5}, and \ref{Fig4}, which illustrate challenging cases identified by our active learning strategies on TableBank-LaTeX and TableBank-Word datasets, respectively. These examples include overlapping tables, complex layouts, and multi-table configurations that traditional random sampling might miss.

Sampling based on table count provides a more significant performance boost for YOLOv9, whereas ambiguity-based sampling methods perform better with CascadeTabNet. This suggests that single-stage detectors benefit significantly from exposure to diverse table configurations, while cascade architectures gain from understanding ambiguity in model predictions. Our experimental results provide strong empirical evidence for the superiority of prediction uncertainty-based and prediction ambiguity-based approaches in table detection tasks, particularly when annotation budgets are constrained. The table count-based active learning consistently offers the best performance-to-annotation-cost ratio, highlighting the importance of incorporating structural information into active learning driven selection criteria.

\section{Conclusion and Future work}

This paper develops the first application of active learning to table detection, presenting sampling strategies for hard examples based on model prediction uncertainty and model prediction ambiguities for table detection. Our experiments across two datasets (TableBank-LaTeX and TableBank-Word) and two state-of-the-art architectures (YOLOv9 and CascadeTabNet) demonstrate that our proposed approach outperforms random sampling across different budgets. 

We plan to extend this work to the more challenging task of table structure recognition, which requires fine-grained detection of cells, rows, and columns within tables. 
The success of this work in table detection suggests its potential value in reducing annotation efforts for the more complex task of structure recognition, where labeling costs are significantly higher due to the detailed annotations required.

%
%
%
\bibliographystyle{splncs04}
\bibliography{refs}
\end{document}